\documentclass[letterpaper]{article} 
\usepackage[]{aaai23}  
\usepackage{amsmath,amsfonts}
\usepackage{array}
\usepackage{times}  
\usepackage{helvet}  
\usepackage{courier}  
\usepackage[hyphens]{url}  
\usepackage{graphicx} 
\usepackage{natbib}  
\usepackage{caption} 
\usepackage{subfig}
\usepackage{booktabs}
\usepackage{xcolor}

\usepackage{amsmath,amssymb} 

\usepackage{algorithm}
\usepackage{algorithmic}

\usepackage{newfloat}
\usepackage{listings}
\DeclareCaptionStyle{ruled}{labelfont=normalfont,labelsep=colon,strut=off} 
\floatstyle{ruled}
\newfloat{listing}{tb}{lst}{}
\floatname{listing}{Listing}
\title{IT-RUDA: Information Theory Assisted Robust Unsupervised Domain Adaptation}
\author{
    Shima Rashidi\textsuperscript{\rm 1},
Ruwan Tennakoon\textsuperscript{\rm 1}, 
Aref Miri Rekavandi\textsuperscript{\rm 2},\\
Papangkorn Jessadatavornwong\textsuperscript{\rm 1},
Amanda Freis\textsuperscript{\rm 3}, 
Garret Huff\textsuperscript{\rm 3}, 
Mark Easton\textsuperscript{\rm 1},
Adrian Mouritz\textsuperscript{\rm 1}, 
Reza Hoseinnezhad\equalcontrib\textsuperscript{\rm 1}, 
Alireza Bab-Hadiashar\equalcontrib\textsuperscript{\rm 1},
}
\affiliations{
    \textsuperscript{\rm 1}RMIT University, Melbourne, Australia\\
        \textsuperscript{\rm 2}The University of Melbourne, Melbourne, Australia\\
        
           \textsuperscript{\rm 3}Ford Motor Company, USA

}

\usepackage{bibentry}

\begin{document}

\maketitle

\begin{abstract}
Distribution shift between train (source) and test (target) datasets is a common problem encountered in machine learning applications. One approach to resolve this issue is to use the Unsupervised Domain Adaptation (UDA) technique that carries out knowledge transfer from a label-rich source domain to an unlabeled target domain. Outliers that exist in either source or target datasets can introduce additional challenges when using UDA in practice. In this paper, $\alpha$-divergence is used as a measure to minimize the discrepancy between the source and target distributions while inheriting robustness, adjustable with a single parameter $\alpha$, as the prominent feature of this measure. Here, it is shown that the other well-known divergence-based UDA techniques can be derived as special cases of the proposed method. Furthermore, a theoretical upper bound is derived for the loss in the target domain in terms of the source loss and the initial $\alpha$-divergence between the two domains. The robustness of the proposed method is validated through testing on several benchmarked datasets in open-set and partial UDA setups where extra classes existing in target and source datasets are considered as outliers. 
\end{abstract}

\section{Introduction}

There is increasing interest in the idea of domain adaptation as it provides a solution for real-world problems where the training and test data do not necessarily have the same distributions \cite{wang2018deep,wilson2020survey}. In particular, closed-set unsupervised domain adaptation (UDA) tackles the machine learning problem where the labeled training (called source) and unlabeled test (called target) datasets are sampled from the same classes but shifted domains (e.g. synthetic vs real-world images or painting vs photographs). Such a domain shift contradicts the machine learning assumption that the marginal distributions of source and target domains are aligned \cite{ben2010theory}. As a result, the accuracy of a model solely trained on the source dataset often drops significantly when tested on the target dataset. This problem has received considerable attention in recent years~\cite{long2018conditional,ma2019gcan,nguyen2021kl,shen2018wasserstein}. 

The problem of unsupervised domain adaptation gets more complicated if outliers exist in either the target or source domains. The outliers can negatively affect the performance of the trained model due to the closed-set assumption of machine learning solutions, especially deep learning models. These types of problems are usually addressed under the umbrella of open-set domain adaptation, called OSDA, \cite{panareda2017open} (outliers existing in the target domain as extra classes private to that domain) and partial domain adaptation, called PDA, \cite{cao2018partial} (outliers existing in the source domain as extra classes private to that domain) in the literature. Many domain adaptation solutions provide complicated algorithms for rejecting unknown target samples \cite{baktashmotlagh2018learning,feng2019attract,gao2020adversarial,saito2018open} or artificially generating them in the source domain to match the two domains \cite{GSOD2022WAC}. A simpler solution, which is somewhat overlooked, is to treat the unknown samples as outliers and apply a robust domain-adaptation method (one example of robust UDA can be found in \cite{balaji2020robust}). The need for a robust method is to mitigate the negative effect of the outliers (private classes) on the domain adaptation process and enable the model to operate on the feature representations of the shared classes unhindered. 

In this paper, a robust domain adaptation method using a general parametric measure from information theory, namely $\alpha$-divergence, is proposed to align the marginal distributions of source and target representation while ignoring private classes (treating them as outliers). Unlike existing methods, which often need a separate network or complicated architectures with some constraints like the 1-Lipschitz constraint on the weights Gradients \cite{balaji2020robust}, our method is simple and can directly estimate the dissimilarity between the two distributions. The benefits of using $\alpha$-divergence are 
    i) The chosen divergence is a general form of several well-known measures such as KL and Reverse KL divergences, tunable via a single parameter $\alpha$. This feature enables one to take advantage of desirable divergence characteristics (like robustness to outliers) by choosing the hyper-parameter $\alpha$.
    ii) It is shown that the proposed loss function is bounded in the target domain in proportion with a function of $\alpha-$divergence of the target and source distributions. In case of perfect alignment of these two distributions, loss (in this paper classification loss) of target and source will be equal, meaning that the network is adapted to the target domain.
    iii) In comparison to previous domain adaptation models, which are mostly limited by running an iteratively trainable separate network to calculate the dissimilarity between source and target samples, the $\alpha$-divergence can be calculated without any additional network or a minimax objective. This leads to a theoretical and efficient metric for the alignment of the two distributions. This is performed by feeding the samples into Gaussian Mixture Models (GMMs) obtained by putting multivariate Gaussian kernels around feature representations of the two domains; i.e. we use the feature embeddings from the encoder as the means of the Gaussians with ones as the variances. With the taken approach, the GMMs are estimated using the neural network directly and separate training of GMMs is no needed. The proposed method is tested on three benchmark datasets: Office31 \cite{saenko2010office31}, VisDA17 \cite{peng2017visda} and Office-Home \cite{venkateswara2017officehome}. The results show that the proposed method outperforms the State Of The Art (SOTA).

\section{Literature review}
Closed-set unsupervised domain adaptation is a well-studied topic in computer vision literature. There are two main streams of work in the literature for addressing the above problem by using deep neural networks, i) Using adversarial networks where a classifier tries to discriminate the target and source samples while a feature-extractor attempts to fool it. As the result, the model finds a representation of the input samples which is indifferent to source and target samples \cite{long2018conditional,ma2019gcan}.  ii) Minimizing the distance or difference of source and target features in the feature space by using distance metrics in the loss function \cite{nguyen2021kl,balaji2020robust}. However, real-world machine learning problems are not always closed-set and unseen classes might exist in either source or target domain. Such problems are addressed as open-set and partial domain adaptation in the literature. 

Open-set domain adaption refers to a situation where the target have unknown samples with different classes than the ones shared with source domain; classified as the class ``unknown''. The concept of open-set models was first presented in \cite{jain2014multi} where Jain et. al. modified the SVMs to reject the samples from unknown classes based on a probability threshold. 
Another stream of works proposed various methods or metrics to separate the unknown classes from known \cite{panareda2017open,baktashmotlagh2018learning,feng2019attract,gao2020adversarial,saito2018open,bucci2020effectiveness}. 
This problem has been approached in multiple ways \cite{liu2021open,fang2020open,pan2020exploring,GSOD2022WAC}. DAOD (distribution alignment with open difference) \cite{fang2020open}, considers the risk of the classifier on unknown classes and tries to regularize it while aligning the distributions. SE-cc \cite{pan2020exploring} applies clustering on the target domain to obtain domain-specific visual cues as additional guidance for the open-set domain adaptation. In \cite{GSOD2022WAC}, the authors tried a different approach where they complemented the source domain via regenerating unknown classes for the source dataset in order to resemble the two datasets.

Partial domain adaptation (PDA) refers to the domain adaptation problem where the source domain contains extra classes which are private to it \cite{cao2018partial}. It was first introduced in \cite{cao2018partial} where the authors used an adversarial network to down-weight the outlier source classes while matching the representations of two domains. Later, example transfer network (ETN) \cite{cao2019learning} was proposed where a transferability weight is assigned to source samples to reduce their negative transfer effect. In deep residual correction network (DRCN) \cite{li2020deep}, a weight-based method is devised to align the target domain with the most relevant source subclasses. BA3US \cite{liang2020balanced} mitigates the imbalance between target and source classes by gradually adding samples from the source to the target dataset.  Adaptive graph adversarial network (AGAN) \cite{kim2021adaptive} uses an adaptive feature propagation technique to utilize the inter- and intra-domain structure and computes the commonness of each sample to be used in the adaptation process.  

It should be noted that although effective, the introduced models mostly suffer from complicated architectures and constraints applied to the optimization process. Here, it is proposed that OSDA and PDA setups can benefit from a robust method which can effectively mitigate the negative transfer effect of unseen classes in either target or source by treating them as outliers. Although interesting, the stream of robust domain adaptation is not pursued in the literature sufficiently. As discussed before, distance-based methods are commonly used to align the distributions of source and target for the purpose of domain adaptation.
Kullback-Leibler divergence \cite{nguyen2021kl} and Wasserstein measure \cite{shen2018wasserstein} have been previously used for this task. Despite their promising results in closed-set domain adaptation scenarios, both measures are sensitive to the influence of outliers. There have been attempts to improve the robustness of the above measures at the cost of adding overhead and increasing the computational cost of training a model \cite{balaji2020robust}. Here, it is proposed to use a more general parametric family of measures called $\alpha$-divergence, which can be tuned by a single parameter $\alpha$ to mitigate the effect of outliers \cite{cichocki2010families}. The benefits of this divergence have been shown in several studies related to robust principal component analysis \cite{rekavandi2020robusta}, robust image processing \cite{rekavandi2021robust,iqbal2019alpha} and robust signal processing \cite{seghouane2019robust,rekavandi2020robust}. To the best of authors' knowledge, the current study is the first attempt to use the $\alpha$-divergence as a robust measure in deep learning based domain adaptation. 
\begin{figure}[!t]
	\centering
	\includegraphics[width=0.78\linewidth]{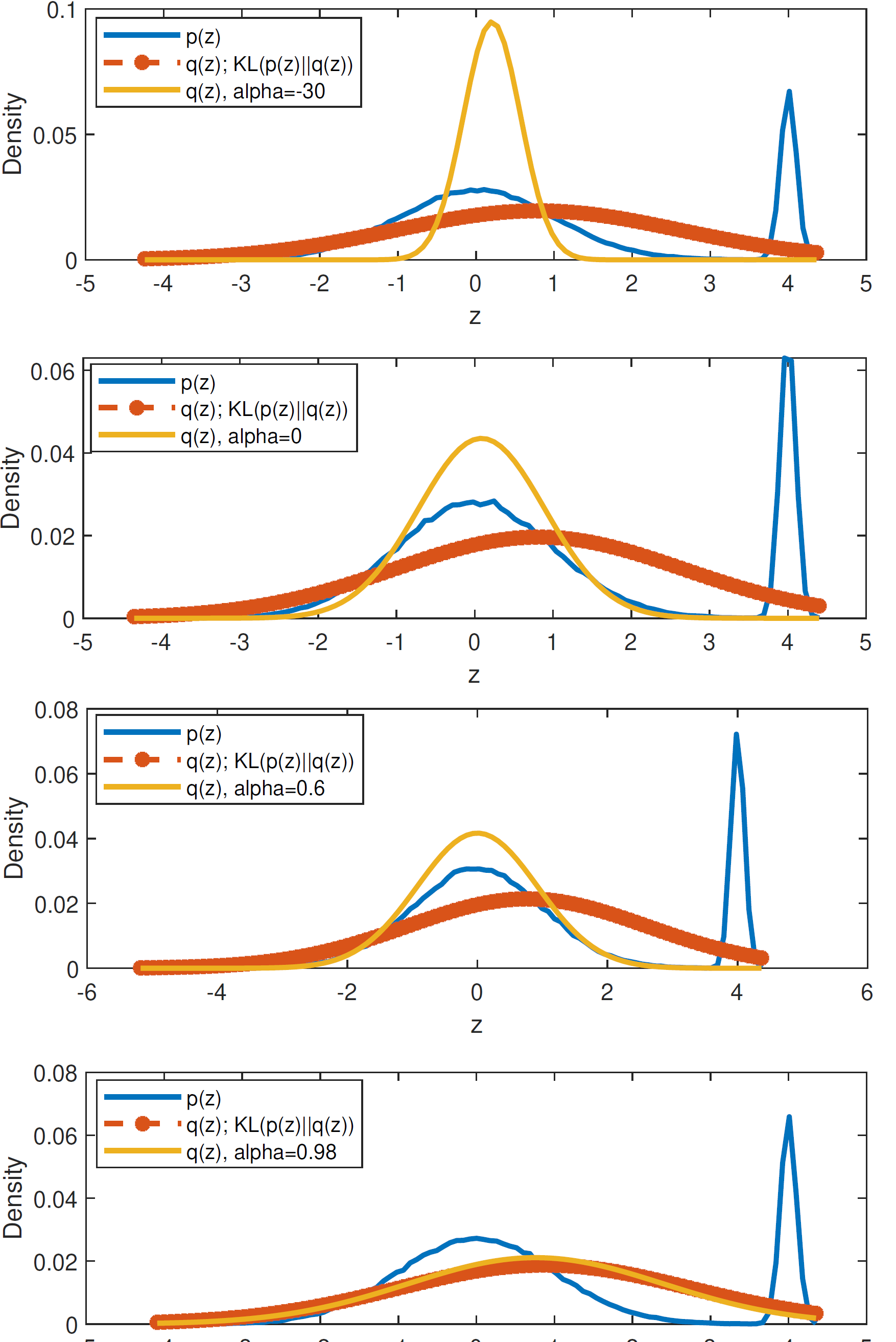}
	
	\caption{Density approximation using $\alpha-$divergence as the measure using different values of hyperparameter $\alpha$.}
	\label{fig1}
\end{figure}
\subsection{Background in $\alpha$-divergence}
The $\alpha$-divergence between two distribution functions, $p(z)$ and $q(z)$, is defined as \cite{cichocki2010families}:\\
\\
$D_{\alpha}(p(z)\|q(z))=\\$
\begin{equation}
    \frac{1}{\alpha(\alpha-1)}\left[\int p(z)^{\alpha} q(z)^{1-\alpha} dz -1 \right].
    \label{eq:alpha_main}
\end{equation}
The tuning parameter $\alpha$ enables the measure to smoothly link the  KL-divergence ($\alpha \rightarrow 1$) to reverse KL-divergence ($\alpha \rightarrow 0$ ) through Hellinger distance ($\alpha \rightarrow 1/2$) \cite{cichocki2010families}. This provides an opportunity to tune the hyperparameter $\alpha$ and inherit the most useful features of this family of measures (e.g. robustness to outliers). This is a non-negative measure that is directly proportional with the dissimilarity of the distributions and would be zero ($D_\alpha =0$) if and only if $p(z)=q(z)$. When $\alpha \rightarrow -\infty$, the estimation of $p(z)$ by $q(z)$ gets exclusive, i.e., $q(z)\leq p(z)$ for all $z$ \cite{cichocki2010families}. This property will be degraded when $\alpha$ tends to $1$, approximating the standard KL divergence. 
The robustness property of this measure is shown in Figure \ref{fig1}. Assume $p(z)$ is an empirical distribution constructed by drawn samples from a linear combination of two Gaussian distributions, e.g., $0.8\mathcal{N}(0,1)$ and $0.2\mathcal{N}(4,0.01)$ where $\mathcal{N}(\mu,\sigma^2)$ is a normal distribution with mean of $\mu$ and variance of $\sigma^2$. It is of interest to estimate $p(z)$ with a single Gaussian density with parameters $\mu$ and $\sigma^2$, i.e., $q(z)=\mathcal{N}(z|\mu,\sigma^2)$. Using $D_{\alpha}$ as a measure, a solution can be found as:  
\begin{eqnarray}
\hat{{\mu}}, \hat{\sigma}^2
&=& 
\arg\min_{\mu, \sigma ^2}D_{\alpha}\left(p(z)\parallel \mathcal{N}(z|\mu, \sigma^2)\right)\nonumber \\
&=&\arg\min_{\mu, \sigma ^2}\frac{1}{\alpha(\alpha-1)}\sum_{i=1}^{N} \big\{ \mathcal{N}(z_i|\mu, \sigma^2) \big \} ^{1-\alpha}.
\end{eqnarray}
The second line is obtained by substituting the $p(z)=\sum_{i=1}^{N} \delta(z-z_i)$ (empirical distribution) in the definition of $D_{\alpha}$ where $\delta(.)$ is the delta-dirac function with the property of $\int_{-\epsilon}^{+\epsilon}\delta(t)dt=1$. 
Figure \ref{fig1} shows the approximation of $p(z)$ by $q(z)$ for different $\alpha$ values ranging from large negative values to 1. As shown in the first plot of Figure \ref{fig1}, for large negative values of $\alpha$, the measure is exclusive and the approximation is tightly around the mass of the actual density $p(z)$ and the second Gaussian component is ignored (considered outlier). However, the variance of the main component is not correctly estimated in this case (an inappropriate setting of $\alpha$). For $\alpha$ between 0 and 1 the robustness property is observed in the second and third plots of figure \ref{fig1}, where the predicted density is a closer approximation of the main component and is less affected by the second component (outlier). Although a better estimation of the variance is achieved in this case, but the mean estimation is a slightly deviated. Finally, when $\alpha \rightarrow 1$ (fourth plot of Figure \ref{fig1}), the KL divergence measure and the $\alpha$-divergence are equivalent and give the same approximation. This approximation is highly affected by the second component and its mean deviates towards it.

\section{Methods}
\subsection{Problem Statement}
In the context of unsupervised domain adaptation, one is given a labeled source dataset $\textbf{X}^s=\{\textbf{x}_j^s,{y}_j^s\}_{j=1}^{N_s} \sim p(\textbf{x},y)$ where $\textbf{x}_j^s\in\mathbb{R}^m$ represents the source samples, ${y}_j^s \in \mathbb{R}$ is its label, $p(\textbf{x},y)$ is the joint data distribution, and $N_s$ is the number of source samples. It is also given an unlabeled dataset from target domain $\textbf{X}^t=\{\textbf{x}_i^t\}_{i=1}^{N_t} \sim q(\textbf{x})$ where $\textbf{x}_i^t\in\mathbb{R}^m$ is a target domain sample, $q(\textbf{x})$ is the target data distribution, and $N_t$ is the number of target samples. In practice, usually $N_s >> N_t$, which is an indication of a knowledge transfer from a large annotated dataset to a small label-free dataset. In domain adaption, the marginal and conditional distributions of the source and target domain are expected to differ, i.e., $p(\textbf{x})\neq q(\textbf{x})$ and $p(y|\textbf{x})\neq q(y|\textbf{x})$. In challenging but more practical cases of domain adaptation such as open-set or partial domain adaptation scenarios, source and target classes are not necessarily the same. In the case of open-set adaptation, ${y}^s$ can be any integer value from the set $\mathcal{Y}_s$, i.e., $\mathcal{Y}_s=\{1,2,\cdots,C\}$ while the unknown label $y^t$ can belong to a more general finite set $\mathcal{Y}_t$, i.e., $\mathcal{Y}_t=\{1,2,\cdots,C,C+1,\cdots,C+K\}$. In the case of partial domain adaptation, the situation is reversed and ${y}^s$ can be any integer value from the set $\mathcal{Y}_s$, i.e., $\mathcal{Y}_s=\{1,2,\cdots,C\}$, while the unknown label $y^t$ belongs to a subset of source labels. In both cases, samples with private labels are considered as outliers and their existence can have a negative transfer effect. 

In this formulation, it is assumed that there is a shared feature extractor parameterized by $\boldsymbol{\theta}$, such that $\textbf{z}=f_\theta(\textbf{x})$ and $f_\theta:\mathbb{R}^m\rightarrow\mathbb{R}^d$, as well as a second shared network to perform the task of interest such as a classification task: $f_\phi:\mathbb{R}^d\rightarrow\mathbb{R}^C$, parameterized by $\boldsymbol{\phi}$. Here $\textbf{z}^t$ and $\textbf{z}^s$ denote the output of the feature extractor (the encoder), i.e., $\textbf{z}^t=f_\theta(\textbf{x}^t)$ and $\textbf{z}^s=f_\theta(\textbf{x}^s)$, respectively. In order to achieve reasonable performance and a successful adaptation to a new domain, domain-invariant techniques aim to determine $f_\theta(.)$ such that shared features along the domains are selected ($\textbf{z}$ is expected to capture common features between two domains). It has been shown that this can be achieved by enforcing the alignment of data representation distributions of the two domains \cite{nguyen2021kl}. In open-set or partial UDA, some samples of the target or source datasets are unseen and can be treated as outliers. In this approach, use of a robust measure would be essential for the representation alignment to mitigate the effect of unseen data samples and increase the chance of developing an adequate feature representation to cover both source and target samples. 
 
\begin{figure*}[!t]
	\centering
	
	\subfloat[]{\includegraphics[width=0.46\linewidth]{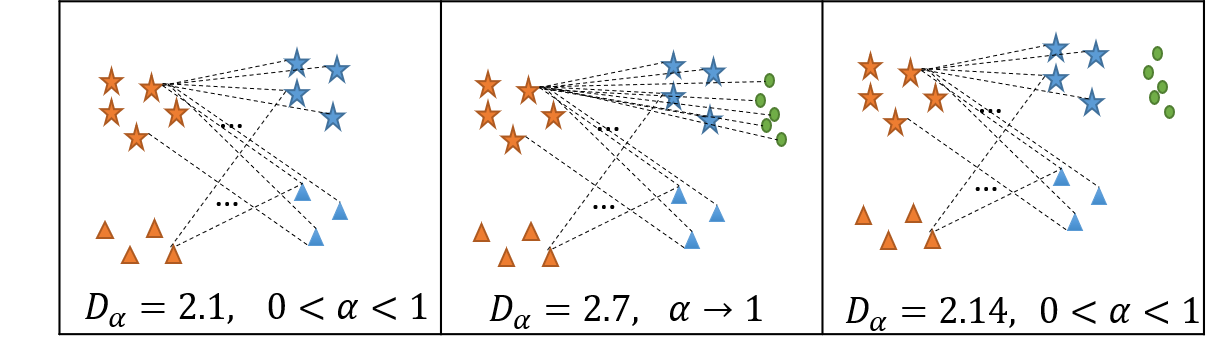}\label{fig:robustness}}
  \hfill
  \subfloat[]{\includegraphics[width=0.4\linewidth]{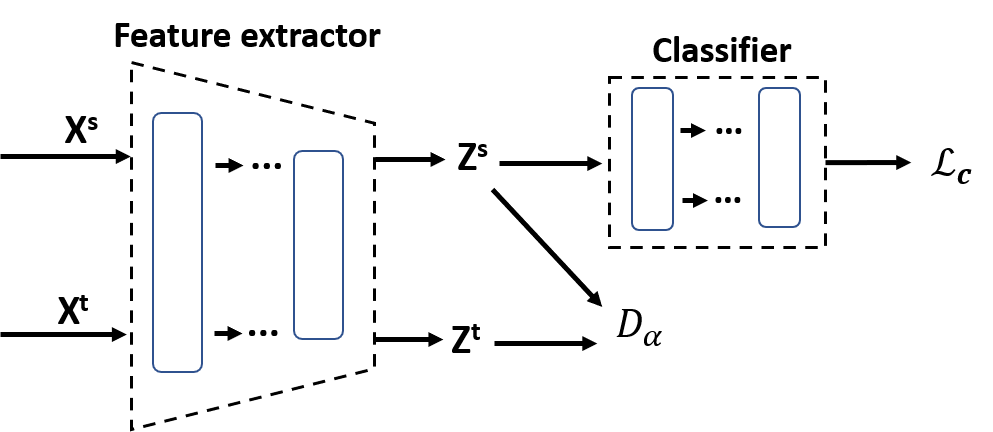}\label{fig:arch}}

	\caption{a) $\alpha$-divergence robustness to outliers. The $\alpha$-divergence of two distributions, left: with two shared classes in a closed-set scenario, middle: two shared classes and one unknown target class when $\alpha$ approaches 1 (equal to KL divergence), right: two shared classes and one unknown target class when $\alpha$ is smaller than one so the unknown samples are treated as outliers and ignored (robust). b) The high-level architecture of the proposed method.}
\end{figure*}
\subsection{$\alpha$-Divergence in Domain Adaptation}

The general architecture of the proposed method for the training phase is shown in Figure \ref{fig:arch}. Both source and target samples are fed into the same network. The intermediate feature representations (\textbf{Z}), just before classifier layer, are computed for both domains. The ultimate goal is to reach a better generalization of the task by extracting intermediate representations from source and target datasets in a way that the representations are invariant to the data domain and have the same distribution. To achieve this, the $D_{\alpha}$ is measured over the distributions of the two domains and then back-propagated as the loss using the Gradient descent. The process adjusts the weights of the feature extractor in a way that it increases the similarity between two distributions. 
The similarity between the two representation sets is defined as: $D_\alpha(q(\textbf{z})||p(\textbf{z}))=D_\alpha(q(f_\theta(\textbf{x}^t))||p(f_\theta(\textbf{x}^s)))$.

It is important to note that the above network can produce a trivial solution when all weights are set to zeros and $\textbf{Z}^s=\textbf{Z}^t=\textbf{0}$, which makes $D_{\alpha}=0$. To avoid this, a source-domain-specific loss function ($\mathcal{L}_c$ Equation \ref{class_loss}) is used to ensure that the extracted features are chosen in a way that the performance of the interested task (i.e. classification) is taken into account. In other words, the network learns parameters of the feature extractor, denoted as $\boldsymbol{\theta}$, as well as the parameters of the classifier $f_\phi$. For the classification task, the loss function can be defined as the negative log likelihood or the cross entropy of the predicted distribution and one hot representation of the labels (Equation \ref{class_loss}).
\begin{equation}\label{class_loss}
    \mathcal{L}_c (x^s,y^s|\theta,\phi)=-\frac{1}{N_s}\sum_{i=1}^{N_s}\sum_{c=1}^{C}\textbf{1}(y_i^s=c)\log f_\phi(f_\theta(x_i^s))_c,
\end{equation}
where $\textbf{1}(y_i^s=c)$ returns 1 only when the argument is correct and zero elsewhere, and $(.)_c$ returns the $c-$th entry of a vector. Note that the loss function of the classifier can be changed for any other task accordingly and this does not affect any other parts of the proposed method. The objective here is to minimize both classification loss and the dissimilarity of the domain distributions (Equation \ref{obj}).
\begin{eqnarray}\label{obj}
\hat{{\theta}}, \hat{\phi}&=& \arg\min_{\theta, \phi } \mathcal{L}_{obj} \nonumber \\
&=& \arg\min_{\theta, \phi } \mathcal{L}_c (x^s,y^s|\theta,\phi) \nonumber \\
&&+ \gamma D_\alpha(q(f_\theta(x^t))||p(f_\theta(x^s))),
\end{eqnarray}
$\gamma$ controls the trade-off between the similarity and source classification loss. As opposed to many other methods in the literature, this method does not need to tune many parameters except $\gamma$ and $\alpha$, where their effect is clear and intuitive. Also, the method does not need any extra networks to estimate the divergence through minimization or maximization tasks as in \cite{balaji2020robust}. 

The theoretical justification for using a $\alpha$-divergence based DA method is as follows. In a general probabilistic case, given the representation $z$, a classifier is trained to predict $y$ through the predictive distribution $\hat{p}(y|z)$, which is an approximation of $p(y|z)$.     

\textbf{Proposition 1:} \textit{If $\alpha^\prime \in (0,1] $, define $\alpha=1-\alpha^\prime $, and assume the loss ($-\log\hat{p}(y|z)$) is bounded by M\footnote{ This is not a restrictive assumption since it can be easily augmented by adding a minimum value to the output probabilities similar to \cite{nguyen2021kl}. }, ${y} \in \mathcal{Y}, {z} \in \mathcal{Z}$, this will result in:}
\begin{flalign}
&l_{target}\leq l_{source} +\frac{M}{\sqrt{2}}\bigg\{\frac{1}{\alpha (\alpha -1)\log e}\bigg\}^{1/2}\nonumber\\
&\quad\quad\times\sqrt{\log\bigg\{1-\alpha (1-\alpha )D_{\alpha}(q(z,y)||p(z,y))\bigg\}}
\end{flalign}
where $l_{source}=\mathbb{E}_{x,y \sim p(x,y), z\sim p(z|x)}[-\log\hat{p}(y|z)]$ and $l_{target}=\mathbb{E}_{x,y \sim q(x,y)}[-\log\hat{p}(y|x)]$.\\

\textit{proof:} From \cite{ben2010theory} and \cite{nguyen2021kl}, it becomes:
\begin{equation} \label{mainineq}
l_{target}\leq l_{source}+\frac{M}{2}\int|p(z,y)-q(z,y)|dzdy, 
\end{equation}
where based on definition, $p(z,y)$ and $q(z,y)$ are the source and target joint distributions,  $|.|$ is the absolute value and the term $\int|p(z,y)-q(z,y)|dzdy$ shows the total variation of the two distributions $p(z,y)$ and $q(z,y)$. Using an appropriate inequality which can link the total variation and the $\alpha$-divergence, here, an upper bound for target loss function is calculated. The inequality adopted here links the total variation with Rényi $\alpha$-divergence ($R_{\alpha^\prime} (.||.)$) which is closely related with the $D_{\alpha^\prime} (.||.)$ used in this paper. In other words, if $\alpha^\prime  \in (0,1] $, it is given \cite{gilardoni2010pinsker}:
\begin{eqnarray}\label{ineq}
\frac{\alpha^\prime }{2}\bigg(\int|p(z,y)-q(z,y)|dzdy\bigg)^2\log e \leq \nonumber \\
 R_{\alpha^\prime }(p(z,y)||q(z,y)). \
\end{eqnarray}
The $R_{\alpha^\prime }(p(z)||q(z))$ is defined by $\frac{1}{\alpha^\prime -1}\log \int p(z)^{\alpha^\prime}  q(z)^{1-\alpha^\prime }dz$ and using the definition of $D_\alpha$, these two divergences are related by 
\begin{flalign}\label{def}
& R_{\alpha^\prime }(p(z,y)||q(z,y)) = \nonumber \\
& \frac{1}{\alpha^\prime -1}\log\{1-\alpha^\prime (1-\alpha^\prime ) D_{\alpha^\prime }(p(z,y)||q(z,y))\}. 
\end{flalign}
inputting (\ref{def}) and (\ref{ineq}) into (\ref{mainineq}) gives
\begin{flalign}
&l_{target}\leq l_{source}+\frac{M}{\sqrt{2}}\bigg\{\frac{1}{\alpha^\prime (\alpha^\prime -1)\log e}\bigg\}^{1/2}
\nonumber \\
&\quad \times \sqrt{\log\bigg\{1-\alpha^\prime (1-\alpha^\prime )D_{\alpha^\prime }(p(z,y)||q(z,y))\bigg\}}.
\end{flalign}
In the last step, a change of variable ($\alpha=1-\alpha^\prime$) is used. By definition, $D_{\alpha^\prime }(p(z,y)||q(z,y))=D_{1-\alpha^\prime }(q(z,y)||p(z,y))$ which means by swapping the position of the distributions, the same value can be obtained when $\alpha^\prime $ is set to $1-\alpha^\prime $. This concludes the presented proof.\\
\textbf{Remark 1:} The above result shows that the loss function in the target domain is upper bounded. The bound is directly related to the classification loss function in the source domain as well as the misalignment between source and target distributions through $D_\alpha$. Furthermore, since $D_\alpha$ includes a family of divergence measures, it can provide a more general parametric model of the distribution misalignment.\\
\textbf{Remark 2:} Based on Proposition 1, as the $D_\alpha (q(z,y)||p(z,y))$ gets smaller over iterations, the argument of the $\log$ function tends to 1 causing the second term of the bound to be tightened. This can be interpreted as indirect minimization of $l_{target}$ if $l_{source}$ does not grow. In limit case, when $D_\alpha \rightarrow 0$, the second term vanishes and the domain distributions are perfectly aligned. This means that minimizing the source loss function is equivalent to minimizing the target loss function.\\
\textbf{Corollary 1:} In the limit case, when $\alpha \rightarrow 1$ in Proposition 1,  $D_\alpha(q(z,y)||p(z,y))$ $ \rightarrow D_{KL}(q(z,y)||p(z,y))$ and using the L’Hopital’s rule, 
\begin{equation}
l_{target}\leq l_{source}+\frac{M}{\sqrt{2}}\sqrt{D_{KL}(q(z,y)||p(z,y))}.
\end{equation}
This is the bound shown in \cite{nguyen2021kl}, which is a special case of the proposed bound. The other related divergence-based upper bounds can also be easily derived by setting the parameter $\alpha$ to the appropriate value. Intuitively, since $p(z,y)$ and $q(z,y)$ use the same discriminator network, if the marginal distributions, $p(z)$ and $q(z)$, are similar, the joint distributions $p(z,y)$ and $q(z,y)$ will be aligned too.

\subsection{Optimization}
In practice, the learning process and the loss calculation are performed in mini-batches and there is no exact parametric model for distributions of the feature representations (\textbf{Z}). In order to make the calculations feasible, within each mini-batch, the source and target distributions are approximated by a mixture of multivariate Gaussian distributions as in \cite{nguyen2021kl}, but with a fixed variance, i.e., for each input $\textbf{x}$, $p(\textbf{z}|\textbf{x})=\mathcal{N}(\textbf{z};\boldsymbol{\mu}(\textbf{x}),{\sigma}^2\textbf{I})$. Finally, given $N_b$ samples from each domain, source and target distributions can be approximated as: $p(\textbf{z})\approx \frac{1}{N_b}\sum_{i=1}^{N_b}p(\textbf{z}|\textbf{x}_i^s)$ and $q(\textbf{z})\approx \frac{1}{N_b}\sum_{i=1}^{N_b}p(\textbf{z}|\textbf{x}_i^t)$, respectively.\\

Inserting the above approximations into the main objective function (\ref{obj}) gives
\begin{eqnarray}\label{final}
\hat{{\theta}}, \hat{\phi}&=& \arg\min_{\theta, \phi } \mathcal{L}_{obj} \nonumber \\
&=& 
\arg\min_{\theta, \phi } \mathcal{L}_c (\textbf{x}^s,y^s|\theta,\phi) +\gamma D_\alpha(q(\textbf{z})||p(\textbf{z})),\nonumber \\
&\approx& \arg\min_{\theta, \phi } \mathcal{L}_c (\textbf{x}^s,y^s|\theta,\phi)  \\ &&+  \frac{\gamma}{\alpha(\alpha-1)}\bigg[\frac{1}{N_b}\sum_{i=1}^{N_b}\bigg\{\frac{p(f_\theta(\textbf{x}_i^t))}{q(f_\theta(\textbf{x}_i^t))}\bigg\}^{1-\alpha}-1\bigg] , \nonumber
\end{eqnarray}
where the last line is obtained from an approximated form of $\alpha-$divergence, i.e., using $\int q^\alpha p^{1-\alpha}=\int q ({p}/{q})^{1-\alpha}=\mathbb{E}_{ q}[({p}/{q})^{1-\alpha}]\approx \frac{1}{B}\sum_{i=1}^{N_b}({p}/{q})^{1-\alpha}$. The $\mathcal{L}_c (\textbf{x}^s,y^s|\theta,\phi)$ is defined as cross entropy ($-\frac{1}{N_b}\sum_{i=1}^{B}\sum_{c=1}^{C}\textbf{1}(y_i^s=c)\log f_\phi(f_\theta(\textbf{x}_i^s))_c \nonumber$) in the OSDA and weighted cross entropy (taken from \cite{liang2020balanced}) in the PDA setup.
%

\section{Experiments}
As there is no established benchmarking procedure to compare the performance of unsupervised domain adaptation (UDA) methods for cases with outliers, the issue of robustness was examined using two existing experimental setups: open-set domain adaptation (OSDA) and partial domain adaptation (PDA) where the private classes of the target and source datasets were considered outliers. An extensive set of comparative results with a wide range of the state of the art methods is presented in Tables 1 and 2. 

\paragraph{Datasets:} 
\textbf{Office31} is a dataset of 4,652 images of 31 categories of common office objects in three different domains called Amazon (A), DSLR (D) and Webcam (W). In OSDA and PDA setups, the first ten classes are shared between the source and target domains, and the last ten classes are either private to the target and source, respectively.
\textbf{Office-Home} is a set of 15500 images of 65 classes of daily objects in 4 different domains: Art (A), Clipart (C), Product(P) and Real-world (R). For the OSDA and PDA setups, the first 25 classes in alphabetical order are chosen as shared target and source classes and the rest are the private classes to the target or source, respectively.
\textbf{VisDA17 \cite{peng2017visda}} is large-scale challenging dataset of images with 12 classes in two domains, synthetic and real, each containing 152,397 and 55,388 images respectively. Following the literature \cite{saito2018open}, the first 6 classes in alphabetical order are used as the known set, and the remaining 6 classes as the unknown one.

\paragraph{Implementation Details:}
Pytorch and pre-trained Resnet50 on Imagenet are used as the backbone of the proposed network with one fully connected layer as the classifier. The models are trained on NVIDIA GeForce GTX GPUs with 12 Gb memory. For the \textbf{OSDA setup}, Cross-entropy is used for the calculation of classification loss. The private classes of target are all labeled as one class of ``unknown.'' The accuracy of the model for each class is calculated for the common classes and the average, equivalent to ``OS*'' in \cite{panareda2017open}, is reported. The feature representation dimension was set to 256 for Office-Home and VisDA17 and 16 for Office31 datasets. The learning rate was chosen as 0.1, decreased during the training using a scheduler. Stochastic gradient descent (SGD) is used as the optimizer with a weight decay of 0.0005 and momentum of 0.9. 
For the \textbf{PDA setup}, the publicly available code of \cite{liang2020balanced} is modified by replacing the adversarial network (used for calculation of transfer loss) with the proposed loss of $\alpha$-divergence. Other parts of their experiments remain unchanged. It should be noted that since $\alpha$-divergence is not symmetrical, to make it robust to the private classes in the source domain in the PDA setup, $p$ and $q$ are exchanged in Equation \ref{eq:alpha_main} (called reverse $\alpha$-divergence in this paper). Batch size and gamma (weight of the similarity loss as in Equation 4) are chosen 64 and 0.1 respectively for both setups. Alpha is chosen as 0.9, 0.7 and 0.7 for both setups for the Office31, VisDA17 and Office-Home respectively. Sigma is set to one for all experiments. All hyper-parameters are tuned through cross-validation on the source dataset.
%

\subsection{Results}
The performance of the proposed method on the three noted datasets is compared with the SOTA models as listed in Tables \ref{tab:office-OSDA} and \ref{tab:officehome-OSDA}. Each experiment is repeated three times and the average accuracy is reported. As can be seen from Tables \ref{tab:office-OSDA} and \ref{tab:officehome-OSDA}, the proposed model presents an increase of average performance in all three tested datasets, Office31, VisDA17 and Office-Home, for both OSDA and PDA setups. 

In the OSDA setup, the proposed method outperforms the baseline as well as the state of the art in all domain shifts by a substantial margin, except in W$\rightarrow$A and D$\rightarrow$A cases of the Office31 dataset. The Office31 dataset has a much smaller number of Webcam and DSLR samples in comparison to Amazon, making it difficult to build an accurate distribution of the classes (when used as a source). Equation (\ref{final}) shows that $\alpha$-divergence is calculated as a function of source and target distributions when fed with sample from the target dataset (the dataset containing the outliers). If the number of source samples is very small, the distributions for each source class (a Gaussian mixture distribution with fixed variance) might not be a good representative of it. This incorrectly results in small probabilities of source distribution at the locations of target samples. Oppositely, for the PDA setup, the proposed method  improves upon the BA3US using a reverse $\alpha$-divergence which is fed with samples from the source dataset. Therefore, the comparatively larger size of the Amazon domain (as the source dataset) in comparison to Dslr and Webcam domains results in accuracy decrease for the transfer tasks $A\rightarrow D$ and $A\rightarrow W$ from Office31 dataset. Furthermore, for Office31, improvements are limited as the domain shift is small and outliers have little negative transfer effect.  For Office-Home, an increase of 0.86 can be observed over BA3US which shows the benefit of using a robust divergence measure compared to an adversarial network for partial domain adaptation. The presented method provides competitive results with the SOTA on the VisDA17 dataset as well, with 1.9 and 1.44 accuracy improvement in OSDA and PDA setups respectively.


\begin{table*}[!t]
  \caption{Accuracy on Office31 and VisDA17 \cite{peng2017visda} dataset in the OSDA (OS*) and PDA setup}
  \label{tab:office-OSDA}
  \centering
  \resizebox{\textwidth}{!}{
  \begin{tabular}{lccccccc|c}
    \toprule
    OSDA setup     &  A $\rightarrow$ D  & A $\rightarrow$ W &  D $\rightarrow$ A   &  D $\rightarrow$ W  &  W $\rightarrow$ A  &  W $\rightarrow$ D & Avg  & syn$\rightarrow$real\\
    \hline
    ATI  \cite{panareda2017open}    & 86.6 & 88.9& 79.6 & 95.3 & 81.4 & 98.7 & 88.4 & 59.0 \\
    UAN   \cite{you2019universal}   & 95.6 & 95.5  & 93.5  & 99.8 & \textbf{94.1} & 81.5 & 93.4 & -\\
    STA   \cite{liu2019separate}   & 95.4 & 92.1  & \textbf{94.1} & 97.1 & 92.1 & 96.6 & 94.6 & 63.9\\
    OSBP  \cite{saito2018open}   & 90.5 & 86.8  & 76.1 & 97.7 & 73.0 & 99.1 &  87.2 & 59.2\\
    ROS  \cite{bucci2020effectiveness}   & 87.5 & 88.4  & 74.8 & 99.3 & 69.7 & 100 & 86.6 & -\\
    InheriTune \cite{kundu2020towards} & 97.1 & 93.2&  91.5 &  97.4 & 88.1 &  99.4 & 94.5 & 64.7\\
    \hline
    IT-RUDA(ours)      & \textbf{99.35} \tiny{$\pm$ 0.2} & \textbf{100} \tiny{$\pm$ 0} & 89.8 \tiny{$\pm$ 0.12} & \textbf{100} \tiny{$\pm$ 0} & 92.7 \tiny{$\pm$ 0.09} & \textbf{100}\tiny{$\pm$ 0} & \textbf{96.97} \tiny{$\pm$ 0.12} & \textbf{66.49} \tiny{$\pm$ 0.11}\\
    increase & $3.75\uparrow$ &  $4.5\uparrow$ & $4.3\downarrow$ & $0.2\uparrow$ & $1.4\downarrow$ & 0 & $2.37\uparrow$ & $1.9\uparrow$\\
    \hline
    \hline
    PDA setup     &  A $\rightarrow$ D  & A $\rightarrow$ W &  D $\rightarrow$ A   &  D $\rightarrow$ W  &  W $\rightarrow$ A  &  W $\rightarrow$ D & Avg & syn $\rightarrow$ real\\
    \hline
    PADA \cite{cao2018partial} & 82.17 & 86.54 & 92.69 & 99.32 & 95.41   & 100 & 92.69 &  53.50\\
    ETN \cite{cao2019learning} & 95.03 & 94.52 & 96.21 & 100 & \textbf{96.73} & 100 & 96.73 & -\\
    DRCN \cite{li2020deep} & 86.00 & 88.05 & 95.60 & 100 & 95.80 & 100 & 94.30 & 58.2\\
    AGAN \cite{kim2021adaptive}  & 97.28 & \textbf{100} & 100 & 94.26 & 95.72 & 95.72 & 97.16 & -\\
    BA3US \cite{liang2020balanced} & \textbf{99.36} & 98.98 & 94.82 & 100 & 94.99 & 98.73 & \textbf{97.81} & 69.86\\
    \hline
    IT-RUDA(ours)  & 96.27 \tiny{$\pm$ 0.02} & 97.22 \tiny{$\pm$ 0.04} & \textbf{96.2} \tiny{$\pm$ 0.08} & \textbf{100} \tiny{$\pm$ 0}& 95.78 \tiny{$\pm$ 0.17} & \textbf{100} \tiny{$\pm$ 0}&  97.57 & 71.3 \tiny{$\pm$ 0.01}\\
    increase & 3.09$\downarrow$ & 2.78$\downarrow$ & 0 & 0 & 0.95 $\downarrow$& 0 & 0.23 $\downarrow$ & 1.44 $\uparrow$\\
    \bottomrule
  \end{tabular}}
\end{table*}

\begin{figure*}[!t]
	\centering
	\includegraphics[width=1\linewidth]{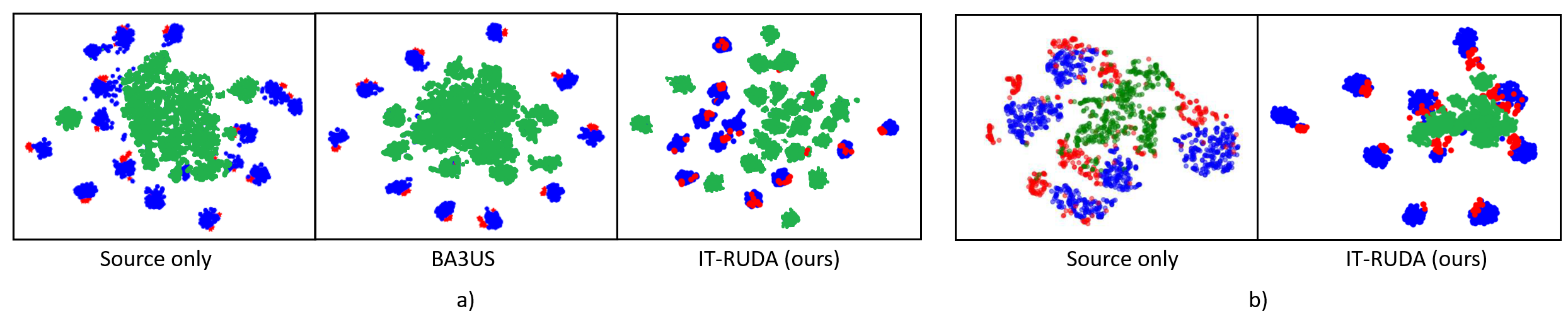}
	\caption{t-SNE visualizations of the feature representations in a) partial UDA task b) open-set UDA task- on Office31 dataset (A $\rightarrow$ D)- \textcolor{blue}{blue: source}, \textcolor{red}{red: target}, \textcolor{green}{green: outlier}}
	\label{fig:features}
\end{figure*}

\begin{table*}
  \caption{Accuracy on Office-Home dataset in the OSDA (OS*) and PDA setup}
  \label{tab:officehome-OSDA}
  \centering
  \resizebox{\textwidth}{!}{%
  \begin{tabular}{lccccccccccccc}
    \toprule
    OSDA setup     & A $\rightarrow$ C     &  A $\rightarrow$ P  &  A $\rightarrow$ R   &  C $\rightarrow$ A  &  C $\rightarrow$ P  &  C $\rightarrow$ R & P $\rightarrow$ A     &  P $\rightarrow$ C  &  P $\rightarrow$ R   &  R $\rightarrow$ A  &  R $\rightarrow$ C  &  R $\rightarrow$ P &  Avg\\
    \hline
    ATI \cite{panareda2017open} & 54.2 & 70.4 & 78.1 & 59.1 & 68.3 & 75.3 & 62.6 & 54.1 & 81.1 & 70.8 & 55.4 & 79.4 & 68.4\\
    OSBP \cite{saito2018open}  & 57.2 & 77.8 &  85.4 & 65.9 & 71.3 & 77.2 & 65.3 & 48.7 & 81.6 & 73.5 & 55.3 & 81.9 & 70.1\\
    ROS \cite{bucci2020effectiveness} & 50.6 & 68.4 & 75.8 & 53.6 & 59.8 & 65.3 & 57.3 & 46.5 & 70.8 & 67.0 & 51.5 & 72.0 & 61.6\\
    PGL \cite{luo2020progressive} & 51.1 & 63.2  & 84.1 & 60.7 & 63.1 & 73.9 & 59.7 & 44.9 & 76.5 & 73.3 & 50.6 & 77.7 & 64.9\\
    GSOD \cite{GSOD2022WAC}  & 58.6 & \textbf{80.5} & 86.5 & 67.2 & \textbf{71.7} & 77.6 & 69.1 & \textbf{54.5} & 82.8 & 77.5 &  \textbf{63.4} & 83.2 & 72.7\\
    \hline
    \shortstack{IT-RUDA(ours) \\ \tiny{ }}   & \shortstack{\textbf{59.32} \\ \tiny{0.21}}&
    \shortstack{79.21 \\ \tiny{0.01}}&
    \shortstack{\textbf{89.23} \\ \tiny{0.2}}&
    \shortstack{\textbf{70.98} \\ \tiny{0.31}}&
    \shortstack{70.75 \\ \tiny{0.14}}&
    \shortstack{\textbf{79.51} \\ \tiny{0.08}}&
    \shortstack{\textbf{72.54} \\ \tiny{0.05}}&
    \shortstack{52 \\ \tiny{0.1}}&
    \shortstack{\textbf{85.86} \\ \tiny{0.19}}&
    \shortstack{\textbf{79.3} \\ \tiny{0.03}}&
    \shortstack{61.1 \\ \tiny{0.01}}& 
    \shortstack{\textbf{85.88} \\ \tiny{0.15}}&
    \shortstack{\textbf{73.80}\\ \tiny{ }}\\
    increase &  0.72$\uparrow$ & 1.29$\downarrow$ & 3.27$\uparrow$ & 3.78$\uparrow$ & 0.95$\downarrow$ & 2.09$\uparrow$ & 3.44$\uparrow$ & 2.5$\downarrow$ & 3.06$\uparrow$ & 1.8$\uparrow$ & 2.3$\downarrow$ & 2.68$\uparrow$ & 1.1$\uparrow$\\
    \hline
    \hline
    PDA setup     & A $\rightarrow$ C     &  A $\rightarrow$ P  &  A $\rightarrow$ R   &  C $\rightarrow$ A  &  C $\rightarrow$ P  &  C $\rightarrow$ R & P $\rightarrow$ A     &  P $\rightarrow$ C  &  P $\rightarrow$ R   &  R $\rightarrow$ A  &  R $\rightarrow$ C  &  R $\rightarrow$ P &  Avg\\
    \hline
    PADA \cite{cao2018partial} & 51.95 & 67.00 & 78.74 & 52.16 & 53.78 & 59.03 & 52.61 & 43.22 & 78.79 & 73.73 & 56.60 & 77.09 & 62.06 \\ 
    ETN \cite{cao2019learning} & 59.24 & 77.03 & 79.54 & 62.92 & 65.73 & 75.01 & 68.29 & 55.37 & 84.37 & 75.72 & 57.66 & 84.54 & 70.45 \\
    DRCN \cite{li2020deep} & 54.00 & 76.40 & 83.00 & 62.10 & 64.50 & 71.00 & 70.80 & 49.80 & 80.50 & 77.50 & 59.10 & 79.90 & 69.00 \\
    AGAN \cite{kim2021adaptive} & 56.36 & 77.52 & 85.09 & 74.20 & 73.84 & 81.12 & 70.80 & 51.52 & 84.54 & 78.97 & 56.78 & 83.42 & 72.82 \\
    BA3US \cite{liang2020balanced} & \textbf{60.62} & 83.16 & 88.39 & 71.75 & 72.79 & 83.40 & 75.45 & \textbf{61.59} & \textbf{86.53} & \textbf{79.52} & 62.80 & 86.05 & 75.98 \\
    \hline
    \shortstack{IT-RUDA(ours) \\ \tiny{ }} & 
    \shortstack{59.22 \\ \tiny{$\pm$ 0.02} }& 
    \shortstack{\textbf{83.85} \\ \tiny{$\pm$ 0} }&
    \shortstack{\textbf{89.56} \\ \tiny{$\pm$ 0.1}} &
    \shortstack{\textbf{74.66} \\ \tiny{$\pm$ 0.08}} &
    \shortstack{\textbf{78.38} \\ \tiny{$\pm$ 0.04} }&
    \shortstack{\textbf{86.97} \\ \tiny{$\pm$ 0.12}} &
    \shortstack{\textbf{76.31} \\ \tiny{$\pm$ 0.043} }&
    \shortstack{59.31 \\ \tiny{$\pm$ 0.068}}& 
    \shortstack{85.81 \\ \tiny{$\pm$ 0.1}} & 
    \shortstack{78.51 \\ \tiny{$\pm$ 0.01}} &
    \shortstack{\textbf{63.04} \\ \tiny{$\pm$ 0.22}} &
    \shortstack{\textbf{86.22} \\ \tiny{$\pm$ 0.18}} &
    \shortstack{\textbf{76.84} \\ \tiny{ }}\\
    increase & 1.4 $\downarrow$ & 0.69$\uparrow$ & 1.17$\uparrow$ & 0.46$\uparrow$ & 4.54$\uparrow$ & 3.57$\uparrow$ &  0.86$\uparrow$ & 2.28$\downarrow$ & 0.72$\downarrow$ & 1.1$\downarrow$ & 0.24$\uparrow$ & 0.17$\uparrow$ & 0.86$\uparrow$\\
    \bottomrule
  \end{tabular}}
\end{table*}
\paragraph{Feature Visualization:}
The t-SNE plots \cite{hinton2008visualizing} of the feature representation for PDA setups are presented in Figure \ref{fig:features}a, comparing source-only, BA3US and IT-RUDA methods for the transfer task A$\rightarrow$D from Office31 dataset. It can be seen that while the domain shift exists in source-only, BA3US and IT-RUDA mitigated its effect through domain adaptation. Importantly, these figures show that IT-RUDA learns the features in a way that the unknown classes remain distinct. Figure \ref{fig:features}b shows the feature representation of the shared classes in the OSDA setup for the task A$\rightarrow$D. As seen, the alignment of the source and target domains is achieved through the proposed method. The proposed robust method effectively ignores the private classes and does not include them for the domain adaptation.

\begin{figure*}[!t]
	\centering
	\includegraphics[width=1\linewidth]{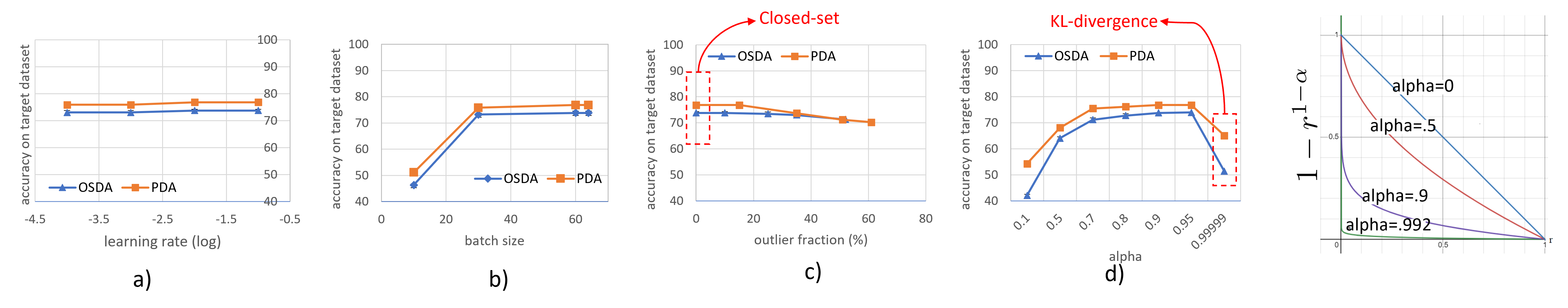}
	\centering

	\caption{Left: The mean of average accuracy reported over target classes for all transfer tasks of the Office-Home dataset in OSDA and PDA setups over 3 runs versus a) log of learning rate, b) batch size of train samples c) outlier-inlier fraction (percentage) d) alpha ($\alpha$). The error bars are standard deviation of accuracy over 3 runs. Right: Changes in the function of $1-r^{1-\alpha}$ for different $\alpha$ values.}
	\label{fig:quality}
\end{figure*}

\paragraph{Sensitivity and Limitations:}
A sensitivity study on learning rate and batch size for both OSDA and PDA setups for the Office-Home dataset was conducted (Figure \ref{fig:quality}a,b). The learning rate does not have significant effect on the reported mean accuracy. However, with small values of batch size the mean accuracy drops noticeably. The proposed method is distribution-based and the samples' distribution are approximated within each batch. As such, using small batch sizes (smaller than 20 samples) can lead to inaccurate distributions. This is considered as a limitation of the proposed method. Furthermore, the effect of outlier-inlier fraction (here the percentage of private classes in all classes) on target classification accuracy for the Office-Home dataset is studied \ref{fig:quality}c. The results show that with the increase of outlier percentage, the negative transfer effect increases and accuracy decreases. However, the accuracy decrease is not considerable, meaning that $\alpha$ is tuned properly for this task and the method is relatively stable.
 
\paragraph{Ablation Studies:}
As discussed in section 3.2, a proper choice of $\alpha$ is needed to mitigate the negative transfer effect of outliers. In Figure \ref{fig:quality}d, mean accuracy of the target dataset for Office-Home dataset for both OSDA and PDA setups is reported with the change of $\alpha$. The results show that with small values of $\alpha$, the estimated divergence is robust and fit around the mass of the distributions. In this case, the method over-reacts and ignores even actual data samples of the distributions. In this case the divergence estimation is non-optimal, resulting in decreased accuracy. With $\alpha$ tending to 1, the measure approximates KL-divergence, which is not robust to outliers, and so accuracy is reduced. It should be noted that the method is not particularly sensitive to the value of alpha within a broad range (0.6 to 0.95); i.e. fine-tuning of $\alpha$ is not necessarily required. 
\paragraph{Fine-Tuning $\alpha$:}
Let's define $r:=p(z)/q(z)$ and so the second term in (\ref{final}) can be written as $\eta\sum_{i=1}^{N_b}\{1-r_i^{1-\alpha}\}$ where $\eta$ is a positive constant. Drawing the function $1-r^{1-\alpha}$ for different $\alpha$ values (Figure \ref{fig:quality})-right shows the sensitivity of the loss function to the outliers. Note that for a normal case where there is no outlier, iterating over $N_b$ target samples, is is expected to obtain all $r$ values close to 1. In contrast, in the case of outlier, this ratio value tends to zero due to absence of any source samples in the feature space neighborhood of target samples. Checking Figure \ref{fig:quality}-right it can be seen that for $\alpha=0.992$ outliers will be problematic, since the gradient of the drawn function has a large value around $r=0$, this makes the weights of encoder to be changed in a direction such that include the outliers in the bulk of distribution mass. Gradually, decreasing $\alpha$ suppresses the Gradient around $r=0$ and pushes the Gradient around $r=1$ upwards which is beneficial for inliers or samples of interest. Now the aim is to make a trade-off between these two cases. It is of interest to bound the Gradient of severe observed outliers in the data. Based on above loss function, the Gradient is $\nabla_r=-\frac{1}{\alpha}r^{-\alpha}$. Let's consider $r=0.01$ as the severe outlier. Now it is only needed to adjust the value of $\alpha$ such that the Gradient at this point ($r=0.01$) be bounded by an arbitrary small value like $\rho$. Hence, the value of $\alpha$ can be determined as a function of threshold value on the $r$ and the bound $\rho$ which may differ application to application and user to user. 

\section {Conclusion}
In this paper, a robust unsupervised dissimilarity-based domain adaptation method using a general measure from information theory, called $\alpha$-divergence, is presented. Use of this measure can, without using complicated networks or optimizations commonly used in OSDA and PDA setups, mitigate the effect of outliers for domain adaptation tasks. The proposed method is tested in OSDA and PDA setups where the private classes are treated as outliers and ignored using a robust divergence measure. A theoretical upper bound of the target domain loss is derived, which shows the source and target domains are aligned; that is, the reduction in classification loss in the source domain leads to reduction of the loss in the target domain as  well. The presented method outperforms the state of the art with an average accuracy of 2.37, 1.9 and 1.1 on Office31, VisDA and Office-Home respectively in the OSDA setup, and an average accuracy -0.23, 1.44 and 0.86 in the PDA setup. 

\section{Acknowledgments}
This work was supported by the Australian Research Council through an ARC Linkage Project grant (LP190100165) in collaboration with Ford Motor Company.

\bibliography{aaai23}

\end{document}